\documentclass[10pt]{article} %
\usepackage[preprint]{tmlr}

\usepackage{amsmath,amsfonts,bm}

\def\eqref#1{equation~\ref{#1}}

\def\1{\bm{1}}

\DeclareMathAlphabet{\mathsfit}{\encodingdefault}{\sfdefault}{m}{sl}
\SetMathAlphabet{\mathsfit}{bold}{\encodingdefault}{\sfdefault}{bx}{n}

\usepackage{hyperref}
\usepackage{url} 

\newcommand{\ppen}[1]{\ensuremath{\mathbf{p}^\text{pen}_{#1}}}
\newcommand{\pearly}[1]{\ensuremath{\mathbf{p}^{\text{early}}_{#1}}}

\newcommand{\Ppen}{\ppen{y=1,\cdots,C}}
\newcommand{\Pearly}{\pearly{y=1,\cdots,C}}

\newcommand{\pnoise}{\mathbf{p}^\text{pen}_\text{ood}}
\usepackage{svg}

\usepackage{multirow} 
\usepackage{acronym}
\usepackage{booktabs}
\acrodef{GROOD}{GRadient-aware Out-Of-Distribution detection}
\acrodef{SNIP}{single-shot network pruning}
\acrodef{CNN}{convolutional neural network}
\acrodef{AUROC}{area under the receiver operating characteristic curve}
\acrodef{DNN}{deep neural network}
\acrodef{KNN}{K-nearest neighbor} 
\acrodef{ID}{in-distribution}
\acrodef{OOD}{out-of-distribution}
\acrodef{ERM}{empirical risk minimization}
\acrodef{iid}{independent and identically distributed}
\acrodef{SSL}{self supervised learning}
\acrodef{PCA}{principle component analysis}
\acrodef{ODIN}{out-of-distribution detector for neural networks}
\acrodef{NC}{neural collapse}
\acrodef{NCP}{nearest class prototype} 

\usepackage{algorithmicx}
\usepackage[ruled]{algorithm}
\usepackage{algpseudocode}
\usepackage{subcaption}
\usepackage{amsmath}
\usepackage{amsfonts}

\usepackage{cleveref} 
\usepackage{graphicx}
\usepackage{multicol}

\makeatletter
\newcommand{\crefnames}[3]{%
  \@for\next:=#1\do{%
    \expandafter\crefname\expandafter{\next}{#2}{#3}%
  }%
}
\makeatother

\crefnames{part,chapter,section}{\S}{\S\S}
\crefnames{part,chapter,section}{\S}{\S\S}
\crefnames{paragraph}{\P}{\P\P}

\title{GROOD: GRadient-Aware Out-of-Distribution Detection} 

\author{%
 \name Mostafa ElAraby \\
 \email \{elarabim\}@mila.quebec \\
 \addr DIRO, Université de Montréal \\
 \addr Mila - Quebec AI Institute
 \AND
 \name Sabyasachi Sahoo \\
 \addr Université Laval, IID \\
 \addr Mila - Quebec AI Institute
 \AND
 \name Yann Pequignot \\
\addr Institut Intelligence et Données (IID)\\
 \addr Université Laval
 \AND
 \name Paul Novello \\
 \addr DEEL, IRT Saint Exupéry
 \AND
 \name Liam Paull \\
 \addr DIRO, Université de Montréal \\
 \addr Mila - Quebec AI Institute \\
 \addr CIFAR AI Chair
}

\def\month{MM}  %
\def\year{YYYY} %
\def\openreview{\url{https://openreview.net/forum?id=XXXX}} %

\begin{document}

\maketitle

\begin{abstract}
Out-of-distribution (OOD) detection is crucial for ensuring the reliability of deep learning models in real-world applications. Existing methods typically focus on feature representations or output-space analysis, often assuming a distribution over these spaces or leveraging gradient norms with respect to model parameters. However, these approaches struggle to distinguish near-OOD samples and often require extensive hyper-parameter tuning, limiting their practicality.
In this work, we propose GRadient-aware Out-Of-Distribution detection (GROOD), a method that derives an OOD prototype from synthetic samples and computes class prototypes directly from In-distribution (ID) training data. By analyzing the gradients of a nearest-class-prototype loss function concerning an artificial OOD prototype, our approach achieves a clear separation between in-distribution and OOD samples.
Experimental evaluations demonstrate that gradients computed from the OOD prototype enhance the distinction between ID and OOD data, surpassing established baselines in robustness, particularly on ImageNet-1k. These findings highlight the potential of gradient-based methods and prototype-driven approaches in advancing OOD detection within deep neural networks.

\end{abstract}

\section{Introduction}\label{sec:intro}
\Acfp{DNN}  have demonstrated exceptional performance across domains such as computer vision, natural language processing, and robotics~\citep{bengio2017deep, lecun2015deep}. Their success largely relies on the assumption that training and test data follow an \ac{iid} pattern~\citep{krizhevsky2012imagenet, simonyan2014very}. However, this assumption often fails in real-world scenarios, where \acp{DNN} encounter \ac{OOD} inputs that deviate significantly from the training distribution~\citep{hendrycks2016baseline}. As a result, models that perform well on \ac{ID} data frequently produce overly confident yet incorrect predictions on \ac{OOD} samples, posing significant risks to safety-critical applications such as healthcare and autonomous driving~\citep{litjens2017survey, bojarski2016end}.
In such scenarios, it becomes imperative for the model to exhibit self-awareness about its own limitations~\citep{mcdropout16icml}. Conventional approaches that focus solely on minimizing the training loss are often ill-equipped to cope with \ac{OOD} samples, thereby jeopardizing the safe and reliable deployment of deep learning systems~\citep{duchi2018learning, arjovsky2019invariant, shen2020stable, liu2021heterogeneous}. 

Consequently, several active lines of research work toward equipping \acp{DNN} with the capability to effectively detect unknown or \ac{OOD} samples. 
Among these, post-hoc \ac{OOD} detection methods stand out as the most convenient, as they utilize the representations of a pre-trained \acp{DNN}, require no additional training, and can be applied to any neural network.
Experimental studies using large benchmarks have underscored the effectiveness of post-hoc \ac{OOD} detection methods, which has led to the development of several specialized \ac{OOD} libraries \citep{yang2022openood,zhang2023openood, kirchheim2022pytorch,oodeel}. 

Existing \ac{OOD} detection methods typically focus on either feature-space-based measures, which assess the distance of inputs to learned feature representations~\citep{sun2022knnood,mahananobis18nips}, or gradient information~\citep{lee2022gradient,sun2022gradient,lee2020gradients, chen2023gaia}, which analyze the model's gradient space. However, these methods often struggle in scenarios where \ac{OOD} samples lie near class boundaries or exhibit characteristics similar to hard in-distribution \ac{ID} examples. Furthermore, current approaches rarely exploit the inherent geometry of learned feature manifolds, limiting their robustness and generalization.

To address these challenges, we propose a novel post-hoc method called \ac{GROOD} that relies on feature and gradient spaces for improved \ac{OOD} discrimination addressing three persistent challenges. First, it improves the detection of near-OOD samples those that are semantically close to the in-distribution by leveraging gradient vectors with respect to an artificial OOD prototype, which provide a more discriminative signal than feature or logit-based scores. Second, GROOD generalizes robustly across diverse architectures, including ResNets and Vision Transformers, where many existing methods suffer performance degradation. Third, it significantly reduces hyper-parameter sensitivity and exhibits stable AUROC performance across training epochs, alleviating the common issue of checkpoint instability. These properties make GROOD a practical and reliable post-hoc framework for real-world deployment.

Our method is inspired by two complementary observations. First, the \ac{NC} property~\citep{papyan2020prevalence} suggests that, at the end of neural network training, within-class variability tends to zero for sample representations in the feature space. This motivates the use of \acf{NCP} classification, which relies on distances to class prototypes, defined as the means of the samples of each class in the feature space. To further enhance the discriminative power of this approach for out-of-distribution detection, we extend the logits to also incorporate distances to an additional \ac{OOD} prototype, alongside the distances to the class prototypes (\cref{subsec:gradient_discrepancy}).

Second, we observe that \ac{OOD} samples tend to exhibit a more dispersed distribution in the feature space compared to \ac{ID} samples. Capitalizing on this characteristic, we introduce an artificial \ac{OOD} prototype, strategically positioned to be distinct from the \ac{ID} class prototypes. By then examining how sample representations respond to this \ac{OOD} prototype, specifically by analyzing gradients of the NCP loss with respect to it, we can gain a more nuanced understanding of the differences between \ac{ID} and \ac{OOD} samples, leading to more effective discrimination in the gradient space.

Our approach differs from traditional post-hoc methods by computing gradients with respect to an artificial \ac{OOD} prototype rather than the network's parameters. The magnitude of these gradients serves as a key indicator: for \ac{ID} data, the \ac{OOD} prototype has a relatively small influence on the confidence of the prediction in the feature space, reflecting stable classification. In contrast, for \ac{OOD} data, the \ac{OOD} prototype has a more substantial influence on the confidence of the prediction; meaning that a smaller shift in the OOD prototype's representation is sufficient to cause a larger change in the classification confidence.
We conduct an extensive empirical study following the recent methodology introduced in the OpenOOD Benchmark~\citep{zhang2023openood}, but we also evaluate our method on other recent architectures. 

Our key results and contributions are summarized as follows.
\begin{itemize}
    \item We propose \ac{GROOD}, a gradient-aware \ac{OOD} detection framework that integrates neural collapse geometry, gradient-space analysis, and synthetic \ac{OOD} generation for robust \ac{OOD} discrimination.

    \item We demonstrate, via an oracle experiment, that an idealized \ac{OOD} prototype significantly improves \ac{OOD} detection.
    
    \item We introduce a novel mixup-based approach for generating synthetic \ac{OOD} data, enhancing \ac{ID}/\ac{OOD} decision boundaries and reducing the need for additional auxiliary \ac{OOD} data. 
    \item We conduct extensive empirical evaluations and ablations, demonstrating \ac{GROOD}'s effectiveness and providing new insights into the interplay of feature and gradient spaces for \ac{OOD} detection.
\end{itemize}

While \ac{NC} is often seen as a limitation for \ac{OOD} detection, as tightly clustered \ac{ID} features can cause overconfident misclassifications, \textsc{GROOD} turns this into a strength. Exploiting the geometric regularity of class prototypes, it detects \ac{OOD} samples not via raw confidence but through their abnormal gradient sensitivity to a fixed synthetic \ac{OOD} prototype. This prototype does not aim to represent the full diversity of \ac{OOD} data, but serves as a consistent reference point, inducing discriminative gradient responses that separate \ac{ID} and \ac{OOD} samples while avoiding the typical overconfidence failure mode.

\section{Related Work}

\paragraph{Neural Network Properties}
Prior research has emphasized the significance of linear interpolation within manifold spaces, with applications ranging from word embeddings~\citep{mikolov2013efficient} to machine translation~\citep{hassan2017synthetic}. Extending these concepts, \citet{verma2019manifold} proposed manifold mixup, a method that smooths decision boundaries and reduces overconfidence near \ac{ID} data.
Our work primarily leverages the \ac{NC} property~\citep{papyan2020prevalence,kothapalli2022neural} and prototype/centroid-based classification. Specifically, we exploit the sensitivity of the \ac{OOD} prototype to enhance the distinction between \ac{ID} and \ac{OOD} samples.  To construct this \ac{OOD} prototype, we employ manifold mixup to generate a synthetic \ac{OOD} dataset, enabling a more robust and structured detection framework.

\paragraph{History of \ac{OOD} Detection}
The study of handling \ac{OOD}  samples has a long history, dating back to early works on classification with rejection~\citep{chow1970optimum, fumera2002support}. These early methods introduced the idea of abstaining from classification when confidence was low, often using simple model families such as SVM~\citep{cortes1995support}. The phenomenon of neural networks producing overconfident predictions on \ac{OOD} data was first revealed by \citet{nguyen2015deep}, highlighting the need for robust detection mechanisms in modern deep learning systems. 
Building on this foundational work, subsequent research has focused on various techniques for detecting \ac{OOD} samples, which can be broadly categorized as output-based, feature-based, and gradient-based methods.

\paragraph{Output-Based Methods}
Many \ac{OOD} detection approaches directly utilize the model’s outputs. Maximum softmax probability, often scaled for calibration, is a classic \ac{OOD} detection metric~\citep{hendrycks2016baseline, guo2017calibration}. 
Building on this, temperature scaling combined with input perturbations has shown promise in refining the separation between \ac{ID} and \ac{OOD} data~\citep{odin18iclr}. Additionally, logits themselves have been used for \ac{OOD} detection, with some methods applying metrics such as KL divergence~\citep{species22icml}.
Beyond these, energy-based methods compute \ac{OOD} scores using energy derived from logits~\citep{energyood20nips}. Refinements such as truncating activations~\citep{sun2021tone, sun2021dice} or removing dominant singular values~\citep{song2022rankfeat, djurisic2023extremely} have been proposed to reduce overconfidence. Generalized entropy scores over logits have also emerged as a robust alternative~\citep{liu2023gen}.  
Unlike the aforementioned techniques, \ac{GROOD} achieves robust \ac{OOD} detection, even for samples near \ac{ID} boundaries, by combining gradient norms with class prototype-based representations.

\paragraph{Feature-Based Methods}
The feature space of neural networks has been a rich avenue for \ac{OOD} detection. Techniques such as Mahalanobis distance from class centroids~\citep{mahananobis18nips, rmd21arxiv} and Gram matrices of features~\citep{gram20icml} are prominent examples. Additional methods utilize noise prototypes~\citep{huang2021feature}, virtual logits~\citep{haoqi2022vim}, and nearest neighbor distances~\citep{sun2022knnood}. Modern Hopfield networks~\citep{she23iclr} have also been explored for this purpose.
Cosine similarity between test samples and class features~\citep{cosinesim20accv, svae20eccv} has gained traction, with some methods proposing the use of singular vectors for enhanced detection~\citep{onedim21cvpr}. Our approach extends these ideas by incorporating an artificial \ac{OOD} prototype into the feature space, creating a novel gradient-based perspective for \ac{OOD} detection.

\paragraph{Gradient-Based Methods}
Gradient-based methods have gained attention for their ability to capture additional information beyond intermediate layers or network outputs. The seminal ODIN approach introduced input perturbations guided by gradients to enhance \ac{OOD} separation~\citep{godin20cvpr}. Subsequent works explored the use of gradients with respect to network weights to quantify uncertainties~\citep{lee2020gradients, igoe2022usefulgrad, huang2021importance}. Another direction utilizes Mahalanobis distances between input gradients, combined with self-supervised classifiers, to detect \ac{OOD} samples~\citep{sun2022gradient}. 
GradNorm calculates an \ac{OOD} score based on the gradient space of the final layer weights~\citep{huang2021importance}.  Recent methods, like \textsc{GAIA}~\citep{chen2023gaia}, leverage gradient-based attribution abnormalities with respect to the feature space, combining channel-wise features and zero-deflation patterns. In contrast, \ac{GROOD} uniquely focuses on gradients with respect to an artificial \ac{OOD} prototype, capturing subtle differences between \ac{ID} and \ac{OOD} data. This approach enables \ac{GROOD} to improve detection performance by leveraging gradient information in conjunction with prototype-based representations.

\section{Preliminaries and Notation}\label{sec:notation}
We first introduce the foundational problem of \ac{OOD} detection and establish the notations.

\subsection{Context and Notations}

Robust deployment of machine learning models in dynamic real-world environments often requires distinguishing between in-distribution (\ac{ID}) and out-of-distribution (\ac{OOD}) data to ensure reliability and safety.
To formalize this challenge, we consider a supervised classification problem. Let \( X \) denote the input space and \( Y = \{1, 2, \ldots, C\} \) the label space, where each input-output pair \((x, y)\) is sampled from a joint data distribution \( P_{XY} \). The training set \( \mathcal{D}_{\text{in}} = \{(x_i, y_i)\}_{i=1}^{n} \) is assumed to be drawn \ac{iid} from \( P_{XY} \). Let \( P_X \) represent the marginal distribution over \( X \). The marginal distribution of the in-distribution data, denoted as \( P_{\text{in}} \), is assumed to be sampled from \( P_X \).

The neural network \( f: X \to \mathbb{R}^{|Y|} \) is trained on samples from \( P_{XY} \) to produce a logit vector, subsequently used for label prediction. The architecture of \( f \) is decomposed as:
\begin{equation}
  f = f^{\text{clf}} \circ f^{\text{pen}}, \quad \text{where} \quad f^{\text{pen}} = f^{\text{mid}} \circ f^{\text{early}}.
\end{equation} 

Here, \( f^{\text{early}} \) extracts low-level features, \( f^{\text{mid}} \) processes mid-level representations, and \( f^{\text{pen}} \) produces penultimate features. The final classification module \( f^{\text{clf}} \) outputs the predictions.

\subsection{Problem Setting: Out-of-Distribution Detection}
When deploying a machine learning model in practice, the classifier should not only be accurate on \ac{ID} samples but should also identify any \ac{OOD} inputs as ``unknown''.

Formally, \ac{OOD} detection can be viewed as a binary classification task. During testing, the task is to determine whether a sample \( x \in X \) comes from \( P_{\text{in}} \) (\ac{ID}) or not (\ac{OOD}). The decision can be framed as a level set estimation:

\[
G_{\tau}(x) =
\begin{cases}
\text{\ac{ID}}, & \text{if } S(x) \leq \tau, \\
\text{\ac{OOD}}, & \text{if } S(x) > \tau,
\end{cases}
\]
where \( S: X \to \mathbb{R} \) is a score function quantifying the likelihood of a sample belonging to the \ac{ID} distribution, and \( \tau \) is a threshold ensuring that a high fraction (e.g., \( 95\% \)) of \ac{ID} data is correctly classified.

\section{GROOD Methodology} \label{sec:methodology}
\begin{figure*}[!ht]
\centering
\includegraphics[height=135pt]{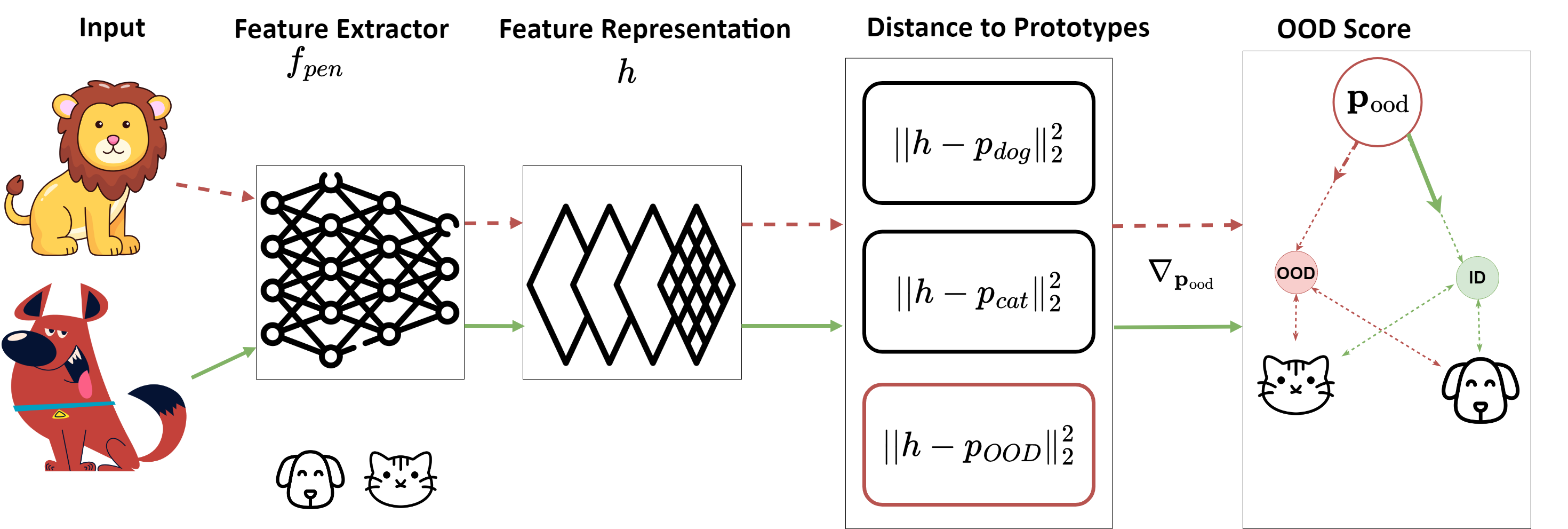}
\caption{Initially, we build \ac{ID} class prototypes as the means of the activations of \ac{ID} data along with an \ac{OOD} prototype capturing \ac{OOD} characteristics (\cref{sec:proto_computation}). Subsequently, gradients of the softmax loss built upon the \acp{NCP} distance as logits are computed w.r.t. \ac{OOD} prototype (\cref{subsec:gradient_discrepancy}).  Finally, the \ac{OOD} score is determined using nearest neighbor distance in the gradient space (\cref{subsec:distance_measurement}). }
\label{fig:grood_overview}
\end{figure*}

\paragraph{Overview}
In this section, we introduce our proposed method \acf{GROOD}, a novel framework for distinguishing between \ac{ID} and \ac{OOD} samples.   
To illustrate the core mechanism, We assume the existence of an OOD prototype (see \Cref{sec:oracle_experiment}), and use it to compute gradients and define our OOD score. 

The method comprises two primary components, illustrated in \Cref{fig:grood_overview}: 
(1) A gradient computation framework that quantifies sample responses to an \ac{OOD} prototype (\cref{subsec:gradient_discrepancy}), and 
(2) A nearest-neighbor scoring mechanism operating in gradient space (\cref{subsec:distance_measurement}).

\subsection{Gradients Computation}\label{subsec:gradient_discrepancy}
Guided by the observations in \Cref{sec:intro}, we build a distance-based classification framework that integrates both class prototypes and an artificial \ac{OOD} prototype. Under the \ac{NC} property~\citep{papyan2020prevalence}, \ac{ID} features concentrate around their class prototypes, making distances a natural choice for logits. In contrast, the \ac{OOD} prototype is positioned away from these clusters and used as a reference point, allowing us to capture the difference between \ac{ID} and \ac{OOD} through their gradients.

For a feature vector $h$ in the penultimate layer space, we define the logit vector as
\begin{equation}
    L(h) = -[\Vert h - \ppen{1} \Vert_2, \ldots, \Vert h - \ppen{C}\Vert_2, \Vert h - \pnoise \Vert_2],
\label{eq:logits}
\end{equation}
where $\ppen{i}$ represents the prototype for class $i$, and $\pnoise$ denotes the \ac{OOD} prototype. These negative distances are transformed into probabilities through the softmax function:
\begin{equation}
    p_i(h) = \frac{\exp(L_i(h))}{\sum_{j=1}^{C+1} \exp(L_j(h))}, \quad i=1,\ldots,C+1,
\label{eq:softmax}
\end{equation}
where $p_{C+1}(h) = p_\text{ood}(h)$ represents the probability of the sample being \ac{OOD}.

This formulation enables us to quantify, through the \ac{NCP} loss, how well a sample aligns with the \ac{ID} prototypes versus the \ac{OOD} prototype. For an \ac{ID} sample, we expect strong alignment with one of the class prototypes and weak alignment with the \ac{OOD} prototype.

For a given feature vector $h$ and a class $y\in{1,..,C,C+1}$, the cross-entropy loss associated with the \ac{NCP} output $[p_i(h)]_{i=1}^{C+1}$ from \eqref{eq:softmax} is given by: 
\begin{equation}
    H(h,y) = -\log p_y(h),
\label{eq:loss}
\end{equation}
The key insight of our method lies in analyzing the gradient of the loss $H(h,y)$ for some (any) \ac{iid} class $y$ with respect to the \ac{OOD} prototype $\pnoise$. Intuitively, this quantity represents the update vector for the \ac{OOD} prototype assuming that the feature $h$ corresponds to an \ac{iid} sample. 
As derived in \cref{ap:grood_derivation}, this gradient can be expressed in closed form as follows:
\begin{equation}
    \nabla H(h) := \nabla_{\pnoise} H(h, C+1) = p_\text{ood}(h) \frac{h - \pnoise}{\|h - \pnoise\|_2}.
\label{eq:grad}
\end{equation}

These computations are grounded in two core observations. The first is inspired by the Neural Collapse phenomenon~\citep{papyan2020prevalence}, which shows that well-trained networks often align penultimate-layer features with their corresponding class prototypes. The second is that \ac{OOD} samples tend to lie outside these tight clusters, often in low-density or dispersed regions. To leverage these properties, we adopt the \(\ell_2\) distance as our metric. This choice is not arbitrary: it ensures that the gradient simplifies to the closed-form in \eqref{eq:grad}, where the gradient's norm directly equals the \ac{OOD} softmax probability  $p_\text{ood}(h)$, and its direction $(h-\pnoise)/\|h-\pnoise\|_2$ encodes the feature’s geometric deviation from the prototype. Such a direct coupling between magnitude, direction, and \ac{OOD} likelihood is essential to \textsc{GROOD}'s mechanism and would not arise with alternative metrics like cosine similarity.

For \ac{ID} samples, we observe smaller gradient norms due to lower \ac{OOD} probabilities. However, the complete gradient vector provides richer information than the norm alone, encoding both magnitude and directional differences between \ac{ID} and \ac{OOD} samples. This allows us to detect \ac{OOD} samples through both the size of the hypothetical update to $\pnoise$ and its direction.

\begin{figure}[tb]
  \centering
  \begin{subfigure}[b]{0.45\textwidth}
    \includegraphics[width=\textwidth]{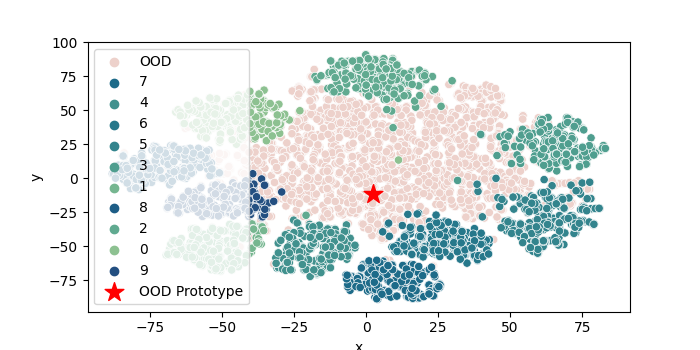}
    \caption{}
  \end{subfigure}
  \begin{subfigure}[b]{0.45\textwidth}
    \includegraphics[width=\textwidth]{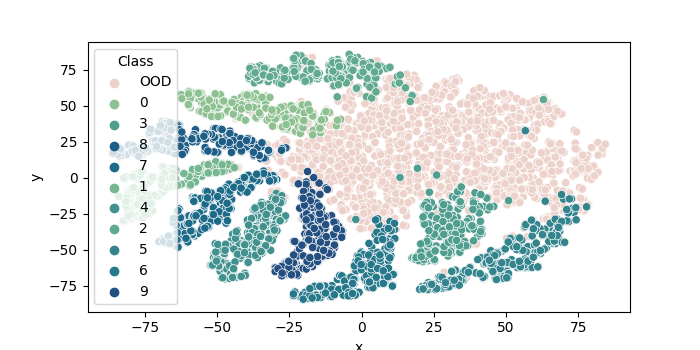}
    \caption{}
  \end{subfigure}
  \caption{t-SNE plots on CIFAR-10 with ResNet-18. (a) Feature space: ID clusters vs. dispersed OOD. (b) Gradient space: clearer ID/OOD separation.}
  \label{fig:tsne_comparison}
\end{figure}

To analyze the potential for enhanced separability, \cref{fig:tsne_comparison} visualizes (a) the penultimate feature space  and (b) the gradient space  using t-SNE.  Our intuition is that the gradient $\nabla H(h)$ will yield a more distinct separation between \ac{ID} and \ac{OOD} samples compared to the feature space. As seen in \cref{fig:tsne_comparison} (a), \ac{ID} samples form class-specific clusters, while \ac{OOD} samples are scatttered in the space. However, the t-SNE plot of the gradient space in \cref{fig:tsne_comparison} (b) reveals a different spatial arrangement, suggesting that the gradient transformation highlights discriminative characteristics for \ac{OOD} detection, which \ac{GROOD} leverages through distance computations in this space (\cref{eq:ood_score_computation}).

\subsection{Final OOD Score Computation}\label{subsec:distance_measurement}

Having established how to compute discriminative gradients with respect to an \ac{OOD} prototype, we now address a key challenge: how to effectively use these gradients to distinguish between \ac{ID} and \ac{OOD} samples. Our solution leverages the observation that \ac{ID} samples produce similar gradient patterns, while \ac{OOD} samples generate distinctly different ones.

\subsubsection{Distance-Based Scoring}
For a test sample $x_{\text{new}}$, we first compute its feature representation $h(x_{\text{new}})$ and corresponding gradient $\nabla H(h(x_{\text{new}}))$. Our OOD score is then defined as the distance to the nearest training gradient: 
\begin{equation} \label{eq:ood_score_computation}
S(x_{\text{new}}) = \min_{x \in \mathcal{D}_{\text{in}}} \| \nabla H(h(x_{\text{new}})) - \nabla H(h(x)) \|_2
\end{equation}

This formulation captures our intuition that \ac{OOD} samples will produce gradients that deviate significantly from those seen during training. The minimum distance provides a natural measure of ``outlierness'' - the further a gradient is from its nearest training neighbor, the more likely the sample is to be \ac{OOD}.

\paragraph{Efficient Implementation}
A naive implementation of nearest neighbor search in high-dimensional gradient space would be computationally prohibitive. We address this challenge using the FAISS library~\cite{faiss}, which provides efficient approximate nearest neighbor search through inverted lists and quantization. This makes our method practical for large-scale applications while maintaining accuracy.

The preceding section outlined the core methodology of \ac{GROOD}, detailing how gradients with respect to an \ac{OOD} prototype are computed and subsequently used within a nearest-neighbor scoring mechanism to distinguish \ac{ID} and \ac{OOD} samples. A critical element underpinning the effectiveness of this methodology is the choice and construction of the \ac{OOD} prototype, $\pnoise$. The nature and location of this prototype in the feature space directly influence the direction and magnitude of the computed gradients, thereby impacting the separability of \ac{ID} and \ac{OOD} samples in the gradient space. Therefore, the following section delves into the specific strategies we employ for \textbf{Prototype Computation} (\cref{sec:proto_computation}), exploring various approaches to define both the class prototypes and, crucially, the \ac{OOD} prototype. These approaches are designed to yield a $\pnoise$ that optimizes the discriminative power of the gradient-based \ac{OOD} score introduced in our methodology.

\section{Prototype Computation} \label{sec:proto_computation}

\subsection{Class and OOD Prototypes} \label{sec:oracle_experiment}

The foundation of our method lies in computing class-discriminative prototypes.
 For each class, we compute prototypes at both early and penultimate layers as the average of feature vectors:
\begin{equation}\label{eq:proto_compute_class}
p_y^l = \frac{1}{|X_y|} \sum_{x \in X_y} f^l(x), \quad l \in \{\text{early}, \text{pen}\}
\end{equation}

where \(X_y\) is the set of training instances in class \(y\).

Similarly, the OOD prototype is computed as the average of feature vectors from a dataset \(X_{ood}\):
\begin{equation}\label{eq:proto_compute_ood}
p_{ood}^\text{pen} = \frac{1}{|X_{ood}|} \sum_{x \in X_{ood}} f^\text{pen}(x) 
\end{equation}
The less variability in the representation space of each ID class as per \ac{NC}\citep{papyan2020prevalence}, the more effective GROOD will be in distinguishing ID and OOD samples.

Clearly, the choice of dataset $X_{ood}$ will have a significant impact on GROOD's ability to differentiate between ID and OOD samples. In the remainder of this section we will first show that, given access to actual OOD data, GROOD performs significantly better than the state of the art (Sec. \ref{subsec:oracle}). Subsequently, we propose a method for synthesizing the OOD data used to calculate the OOD prototype that approximates the performance of the priviliged case (Sec. \ref{sec:practical_proto}).

\subsubsection{``Oracle'' Experiment}
\label{subsec:oracle}
\begin{table}[ht]
\caption{Oracle experiment results compared to SOTA excluding \ac{GROOD} from \Cref{tab:main_results}. AUROC (\%) on far-OOD and near-OOD detection using 100 prototype samples. Results are averaged over different checkpoints; standard deviations in parentheses.}
\label{tab:oracle_results}
\centering
\resizebox{\textwidth}{!}{
\begin{tabular}{llcccccc}
\toprule
\multirow{2}{*}{\textbf{ID Dataset}} & \multirow{2}{*}{\textbf{Architecture}} & \multicolumn{2}{c}{\textbf{Local Oracle}} & \multicolumn{2}{c}{\textbf{Global Oracle}} & \multicolumn{2}{c}{\textbf{SOTA}} \\
\cmidrule(lr){3-8}
& & \multicolumn{6}{c}{\textbf{AUROC (\%)} $\uparrow$ } \\
\cmidrule(lr){3-8}
& & Far-OOD & Near-OOD & Far-OOD & Near-OOD & Far-OOD & Near-OOD \\
\midrule
CIFAR-10 & ResNet-18 & 96.7 (±0.2) &  95.4 (±0.1) & 94.8 (±0.1) & 90.8 (±0.1) & 94.7 & 90.7 \\
CIFAR-100 & ResNet-18 & 94.8 (±0.2) & 85.5 (±0.5) & 88.1 (±0.03) & 82.1 (±0.1)  & 82.4 & 81.3 \\
ImageNet-200 & ResNet-18 & 98.8 (±0.2) & 92.1 (±0.2) & 94.22 (±0.02) & 84.1 (±0.2) & 93.16 & 82.9 \\
ImageNet-1K & ResNet-50 & 97.9 & 91.0 &  96.2 & 79 & 95.1 & 78.1 \\ 
\bottomrule
\end{tabular}
}%
\end{table}

To validate the core idea that an \ac{OOD} prototype can help distinguish \ac{ID} and \ac{OOD} data, we performed an ``oracle'' experiment, assuming temporary access to some \ac{OOD} information.

\textbf{Local Oracle (Idealized):} For each \ac{ID}-\ac{OOD} test pair, we used a small sample (100) from the \textit{test} \ac{OOD} data to build a specific \ac{OOD} prototype. We then tested the remaining \ac{OOD} data. This setup essentially asks: ``If we had perfect knowledge of a small subset of the specific \ac{OOD} data we'd encounter, how well could our method perform?''. The remarkable performance achieved in this setting (over \(95\%\) AUROC on far-\ac{OOD} detection, as shown in \cref{tab:oracle_results}) highlights the inherent potential of an \ac{OOD} prototype tailored to the specific distributional shift.

\textbf{Global Oracle (Generalizable):} For each \ac{ID} dataset, we used a small ``validation'' portion of \textit{all other} \ac{OOD} datasets to create a single, general \ac{OOD} prototype. We then tested on the held-out portion of each specific \ac{OOD} dataset.  This setup aims to mimic a scenario where we have access to some diverse \ac{OOD} data (the validation sets) but not the specific \ac{OOD} data we are currently testing on. The results from the Global Oracle provide insights into how well a more general \ac{OOD} prototype can generalize across different out-of-distribution scenarios.

Despite the strong performance in the far-\ac{OOD} detection tasks under both oracle settings, the notably lower performance on near-\ac{OOD} detection underscores the inherent difficulty of distinguishing between distributions that are semantically or statistically close to the in-distribution data. Nevertheless, the oracle experiments strongly suggest that the concept of an \ac{OOD} prototype holds significant promise for \ac{OOD} detection when a representative prototype can be effectively determined. 

On the other hand, in real-world scenarios, we cannot construct this oracle prototype using test \ac{OOD} data. This raises a key question: \textit{How can we approximate this optimal \ac{OOD} prototype without access to the test distribution?}

\subsection{Practical OOD Prototype Construction} \label{sec:practical_proto}

The oracle experiments presented in \cref{sec:oracle_experiment} demonstrated the significant potential of employing an OOD prototype to distinguish between \ac{ID} and \ac{OOD} data, achieving high performance when even approximate knowledge of the OOD distribution was available. This motivates our goal: to effectively approximate such an optimal OOD prototype, $\pnoise$, without requiring access to the specific test \ac{OOD} distribution, which is unavailable in practical scenarios.

While real-world \ac{OOD} samples exhibit considerable diversity, representing them with a single prototype $\pnoise$ proves effective within our gradient-aware framework. Our approach does not aim to represent all \ac{OOD} samples geometrically, but rather uses the prototype as a fixed reference point. The core \ac{OOD} score relies on the sensitivity of the \ac{NCP} loss to this prototype, measured via the gradient $\nabla_{\pnoise} H(h)$ (\cref{sec:methodology}, \cref{eq:grad}). 
We hypothesize that \ac{OOD} samples, inherently deviating from learned \ac{ID} manifolds, exhibit distinct gradient sensitivity patterns (both magnitude and direction) relative to this \ac{OOD} reference point, allowing separation from more stable \ac{ID} samples.

We propose several complementary approaches to construct $\pnoise$, leveraging information from an auxiliary OOD dataset \(X_{ood}\):

\paragraph{Synthetic OOD Generation using mixup}\label{par:synthetic_data}
Our first approach requires no external OOD data, instead synthesizing OOD-like features by exploiting decision boundaries. We perform guided prototype interpolation towards the second-highest predicted class \(c_2\) at an early layer (after the first block):
\begin{equation}\label{eq:synthetic_data}
\hat{h}(x) = f^{\text{mid}} \left( \lambda f^{\text{early}}(x) + (1-\lambda) \pearly{c_2} \right)
\end{equation}
where \(\lambda=0.5\) positions the synthetic samples near decision boundaries. This approach leverages our observation that early layer representations are more sensitive to perturbations, making them ideal for generating \ac{OOD}-like features.
Although effective, mixup-generated OOD samples interpolate between ID classes and may not fully capture \ac{OOD} diversity.

\paragraph{Auxiliary OOD Validation}
When available, we can utilize a small auxiliary OOD validation set as \(X_{ood}\) to construct the prototype following \cref{eq:proto_compute_ood} using 100 OOD validation samples.
Importantly, we ensure these samples have no category overlap with the test set. Our method shows remarkable stability to the specific choice of validation samples, with a maximum AUROC standard deviation of only 0.5\% across five different random selections.

\paragraph{Proximity-Based OOD Filtering (Postprocessing)}
Initially, we explored constructing the \ac{OOD} prototype by simply averaging feature vectors from all available \ac{OOD} samples. However, this approach yielded prototypes that lacked sufficient discriminative power, resulting in poor separation between \ac{ID} and \ac{OOD} data. To address this, we introduce a proximity-based filtering step to refine the \ac{OOD} prototype, enhancing its ability to distinguish \ac{OOD} samples from \ac{ID} samples.

Specifically, given a set of candidate \ac{OOD} feature vectors, we discard \ac{OOD} samples whose distance \( d_i = \min_{j} \| f^{\text{pen}}(o_i) - p_j \|_2 \) falls below an adaptive threshold \( \tau \), computed as the \( q \)-th quantile of \( \{ d_i \}_{i=1}^{n_{\text{ood}}} \). This filtering step refines the \ac{OOD} prototype by ensuring separation from \ac{ID} data while preserving representativeness.

A comparison of alternative \(X_{OOD}\) generation methods is presented in the \cref{ap:choice_xood}.
The results reported in the remainder of this paper utilize the \ac{OOD} prototype derived from synthetic data generated from \ac{ID} samples.

\section{Experiments}
For a comprehensive evaluation of \ac{GROOD}'s performance, we adhere to the OpenOOD v1.5 criteria~\cite{zhang2023openood, yang2022openood}. Results are aggregated in \cref{tab:main_results} including the performance of the nearest class prototype used instead of the classification head. 
For robustness, each evaluation metric except for ImageNet-1k is derived from three runs with unique initialization seeds. In the case of ImageNet-1k, we report results based on a single seed run provided by torchvision~\cite{torchvision2016}.

\paragraph{Experimental Setup}
We evaluate performance using the Area Under the Receiver Operating Characteristic curve (AUROC), where higher values are better. Our benchmarking strategy follows the OpenOOD framework~\cite{zhang2023openood, yang2022openood}, involving four core \ac{ID} datasets (CIFAR-10, CIFAR-100, ImageNet-200, ImageNet-1k) and examining both near and far-\ac{OOD} scenarios. 
For CIFAR-10/100 (50k train/10k test images each), near-\ac{OOD} datasets are CIFAR-100/TinyImageNet, and far-\ac{OOD} are MNIST, SVHN, Textures, and Places365. 
For ImageNet-200 (200 classes, 64x64 resolution), near-\ac{OOD} datasets are SSB-hard/NINCO, and far-\ac{OOD} are iNaturalist, Textures, and OpenImage-O; ImageNet-1k shares these \ac{OOD} datasets. Regarding configuration, we deploy ResNet-18 for CIFAR-10/100 and ImageNet-200 using pre-trained checkpoints from OpenOOD for consistency, testing with three distinct seeds for robustness. For ImageNet-1k, we apply pre-trained torchvision models (ResNet-50, ViT-B-16, Swin-T) to explore \ac{GROOD}'s effectiveness in a broader context than OpenOOD v1~\cite{yang2022openood}.
To allow for reproducibility and facilitate further research, the complete code, including training and evaluation scripts, is available at: \url{https://github.com/mostafaelaraby/Gradient-Aware-OOD-Detection}.

For CIFAR-10/100 and ImageNet-200, we train ResNet-18 models for 100 epochs using SGD with momentum 0.9, weight decay 5e\text{-}4, cosine learning rate decay (starting from 0.1), and batch sizes of 128 (CIFAR) and 256 (ImageNet-200). For ImageNet-1K, we use pretrained models from \texttt{torchvision} (ResNet-50, ViT, Swin). When fine-tuning is required, we follow the OpenOOD v1.5 protocol~\citep{zhang2023openood} with 30 epochs, learning rate 0.001, and batch size 256.

\begin{table*}[ht]
    \caption{Main results from OpenOOD v1.5 on standard OOD detection (AUROC). \ac{GROOD} using synthetic \ac{OOD} data (\cref{par:synthetic_data}) shows superior results compared to existing baselines.}
    \label{tab:main_results}  
    \centering
    \resizebox{\textwidth}{!}{%
    \begin{tabular}{l|cc|cc|cc|cc} 
        \toprule
         & \multicolumn{2}{c|}{\textbf{CIFAR-10}} & \multicolumn{2}{c|}{\textbf{CIFAR-100}} & \multicolumn{2}{c|}{\textbf{ImageNet-200}} & \multicolumn{2}{c}{\textbf{ImageNet-1K}} \\
        \midrule
        \multicolumn{1}{c|}{ID Acc. (\%)} & \multicolumn{2}{c|}{95.06\%\textsubscript{(\textpm 0.30)}} & \multicolumn{2}{c|}{77.25\%\textsubscript{(\textpm 0.10)}} & \multicolumn{2}{c|}{86.37\%\textsubscript{(\textpm 0.08)}} & \multicolumn{2}{c}{76.18\%} \\ 
         
        \multicolumn{1}{c|}{NCP Acc. (\%)} & \multicolumn{2}{c|}{95.01\%\textsubscript{(\textpm 0.08)}} & \multicolumn{2}{c|}{77.10\%\textsubscript{(\textpm 0.001)}} & \multicolumn{2}{c|}{85.75\%\textsubscript{(\textpm 0.003)}} & \multicolumn{2}{c}{71.38\%} \\
        \midrule
        Method & Near-OOD (\%) $\uparrow$ & Far-OOD (\%) $\uparrow$ & Near-OOD (\%) $\uparrow$ & Far-OOD (\%) $\uparrow$ & Near-OOD (\%) $\uparrow$ & Far-OOD (\%) $\uparrow$ & Near-OOD (\%) $\uparrow$ & Far-OOD (\%) $\uparrow$ \\
        \midrule
        OpenMax \cite{openmax16cvpr} & 87.62\%\textsubscript{(\textpm 0.29)} & 89.62\%\textsubscript{(\textpm 0.19)}  & 76.41\%\textsubscript{(\textpm 0.25)} & 79.48\%\textsubscript{(\textpm 0.41)}  & 80.27\%\textsubscript{(\textpm 0.10)} & 90.20\%\textsubscript{(\textpm 0.17)}  & 74.77\% & 89.26\%  \\
        MSP \cite{hendrycks2016baseline} & 88.03\%\textsubscript{(\textpm 0.25)} & 90.73\%\textsubscript{(\textpm 0.43)}  & 80.27\%\textsubscript{(\textpm 0.11)} & 77.76\%\textsubscript{(\textpm 0.44)}  & 83.34\%\textsubscript{(\textpm 0.06)} & 90.13\%\textsubscript{(\textpm 0.09)}  & 76.02\% & 85.23\%  \\
        ODIN \cite{odin18iclr} & 82.87\%\textsubscript{(\textpm 1.85)} & 87.96\%\textsubscript{(\textpm 0.61)}  & 79.90\%\textsubscript{(\textpm 0.11)} & 79.28\%\textsubscript{(\textpm 0.21)}  & 80.27\%\textsubscript{(\textpm 0.08)} & 91.71\%\textsubscript{(\textpm 0.19)}  & 74.75\% & 89.47\%  \\
        MDS \cite{mahananobis18nips} & 84.20\%\textsubscript{(\textpm 2.40)} & 89.72\%\textsubscript{(\textpm 1.36)}  & 58.69\%\textsubscript{(\textpm 0.09)} & 69.39\%\textsubscript{(\textpm 1.39)}  & 61.93\%\textsubscript{(\textpm 0.51)} & 74.72\%\textsubscript{(\textpm 0.26)}  & 55.44\% & 74.25\%  \\
        EBO \cite{energyood20nips} & 87.58\%\textsubscript{(\textpm 0.46)} & 91.21\%\textsubscript{(\textpm 0.92)}  & 80.91\%\textsubscript{(\textpm 0.08)} & 79.77\%\textsubscript{(\textpm 0.61)}  & 82.50\%\textsubscript{(\textpm 0.05)} & 90.86\%\textsubscript{(\textpm 0.21)}  & 75.89\% & 89.47\%  \\
        ReAct \cite{sun2021tone} & 87.11\%\textsubscript{(\textpm 0.61)} & 90.42\%\textsubscript{(\textpm 1.41)}  & 80.77\%\textsubscript{(\textpm 0.05)} & 80.39\%\textsubscript{(\textpm 0.49)}  & 81.87\%\textsubscript{(\textpm 0.98)} & 92.31\%\textsubscript{(\textpm 0.56)}  & 77.38\% & 93.67\%  \\
        DICE \cite{sun2021dice} & 78.34\%\textsubscript{(\textpm 0.79)} & 84.23\%\textsubscript{(\textpm 1.89)}  & 79.38\%\textsubscript{(\textpm 0.23)} & 80.01\%\textsubscript{(\textpm 0.18)}  & 81.78\%\textsubscript{(\textpm 0.14)} & 90.80\%\textsubscript{(\textpm 0.31)}  & 73.07\% & 90.95\%  \\ 
        GradNorm \cite{huang2021importance} & 54.90\%\textsubscript{(\textpm 0.98)} & 57.55\%\textsubscript{(\textpm 3.22)}  & 70.13\%\textsubscript{(\textpm 0.47)} & 69.14\%\textsubscript{(\textpm 1.05)}  & 72.75\%\textsubscript{(\textpm 0.48)} & 84.26\%\textsubscript{(\textpm 0.87)}  & 72.96\% & 90.25\%  \\
        MLS \cite{species22icml} & 87.52\%\textsubscript{(\textpm 0.47)} & 91.10\%\textsubscript{(\textpm 0.89)}  & 81.05\%\textsubscript{(\textpm 0.07)} & 79.67\%\textsubscript{(\textpm 0.57)}  & 82.90\%\textsubscript{(\textpm 0.04)} & 91.11\%\textsubscript{(\textpm 0.19)}  & 76.46\% & 89.57\%  \\
        VIM \cite{haoqi2022vim} & 88.68\%\textsubscript{(\textpm 0.28)} & 93.48\%\textsubscript{(\textpm 0.24)}  & 74.98\%\textsubscript{(\textpm 0.13)} & 81.70\%\textsubscript{(\textpm 0.62)}  & 78.68\%\textsubscript{(\textpm 0.24)} & 91.26\%\textsubscript{(\textpm 0.19)}  & 72.08\% & 92.68\%  \\
        KNN \cite{sun2022knnood} & 90.64\%\textsubscript{(\textpm 0.20)} & 92.96\%\textsubscript{(\textpm 0.14)}  & 80.18\%\textsubscript{(\textpm 0.15)} & 82.40\%\textsubscript{(\textpm 0.17)}  & 81.57\%\textsubscript{(\textpm 0.17)} & 93.16\%\textsubscript{(\textpm 0.22)}  & 71.10\% & 90.18\%  \\
        SHE \cite{she23iclr} & 81.54\%\textsubscript{(\textpm 0.51)} & 85.32\%\textsubscript{(\textpm 1.43)}  & 78.95\%\textsubscript{(\textpm 0.18)} & 76.92\%\textsubscript{(\textpm 1.16)}  & 80.18\%\textsubscript{(\textpm 0.25)} & 89.81\%\textsubscript{(\textpm 0.61)}  & 73.78\% & 90.92\%  \\
        ASH \cite{djurisic2023extremely} & 75.27\%\textsubscript{(\textpm 1.04)} & 78.49\%\textsubscript{(\textpm 2.58)}  & 78.20\%\textsubscript{(\textpm 0.15)} & 80.58\%\textsubscript{(\textpm 0.66)}  & 82.38\%\textsubscript{(\textpm 0.19)} & 93.90\%\textsubscript{(\textpm 0.27)}  & 78.17\% & 95.1\%  \\
        GAIA \cite{chen2023gaia} & 85.1\%\textsubscript{(\textpm 10.2)} & 92.1\%\textsubscript{(\textpm 2.9)}  & 70.75\%\textsubscript{(\textpm 2.11)} & 86.2\%\textsubscript{(\textpm 5.1)}  & 75.1\%\textsubscript{(\textpm 9.8)} & 88.14\%\textsubscript{(\textpm 1.8)}  & 66.98\% & 90.2\%  \\
        CIDER \cite{ming2023cider} & 90.7\%\textsubscript{(\textpm 0.1)} & \textbf{94.7\%}\textsubscript{(\textpm 0.36)}  & 73.10\%\textsubscript{(\textpm 0.3)} & 80.49\%\textsubscript{(\textpm 0.68)}  & 80.58\%\textsubscript{(\textpm 1.7)} & 90.66\%\textsubscript{(\textpm 1.6)}  & 68.9\% & 92.18\%  \\
        GEN \cite{liu2023gen} & 88.2\%\textsubscript{(\textpm 0.3)} & 91.35\%\textsubscript{(\textpm 0.55)}  & \textbf{81.31\%}\textsubscript{(\textpm 0.1)} & 79.68\%\textsubscript{(\textpm 0.6)}  & 82.9\%\textsubscript{(\textpm 0.34)} & 91.36\%\textsubscript{(\textpm 0.45)}  & 76.85\% & 89.76\%  \\  
        fdbd \cite{liu2023fast} & 90.4\%\textsubscript{(\textpm 0.12)} & 93.16\%\textsubscript{(\textpm 0.25)} & 81.2\%\textsubscript{(\textpm 0.05)} & 79.85\%\textsubscript{(\textpm 0.15)} & \textbf{84.2\%}\textsubscript{(\textpm 0.3)} & 93.4\%\textsubscript{(\textpm 0.2)} & 76.6\% & 92.7\% \\
        NCI \cite{liu2025detecting} & 88.8\%\textsubscript{(\textpm 0.1)} & 91.26\%\textsubscript{(\textpm 0.2)} & 81\%\textsubscript{(\textpm 0.2)} & 81.3\%\textsubscript{(\textpm 0.15)} & 83.5\%\textsubscript{(\textpm 0.4)} & \textbf{93.7\%}\textsubscript{(\textpm 0.15)} & 78.6\% & \textbf{95.5\%} \\
        \textbf{\ac{GROOD} (Ours)} & \textbf{91.16\%}\textsubscript{(\textpm 0.001)} & 93.8\%\textsubscript{(\textpm 0.02)}  & 78.9\%\textsubscript{(\textpm 0.05)} & \textbf{84.44\%}\textsubscript{(\textpm 0.9)}  & \textbf{83.4\%}\textsubscript{(\textpm 0.12)} & 92.19\%\textsubscript{(\textpm 0.12)}  & \textbf{78.91\%} & 94.8 \%  \\
        \bottomrule
        \end{tabular}%
    } %
\end{table*}

\paragraph{Main Results Discussion}
GROOD shows strong performance across datasets, but performance varies depending on the trade-off between Near and Far-OOD detection. On CIFAR-100, GROOD achieves state-of-the-art Far-OOD AUROC (84.44\%) among post-hoc methods and competitive Near-OOD performance (78.9\%). While some methods like VIM~\citep{haoqi2022vim} (81.70\%) report higher Near-OOD scores, they trade off Far-OOD robustness.

On ImageNet-1k, GROOD achieves a top Far-OOD score (94.8\%) but a lower Near-OOD score (78.91\%), whereas CombOOD~\citep{rajasekaran2024combood} yields 95.22\% Near-OOD and 90.24\% Far-OOD.

These results reflect a key characteristic of GROOD: its effectiveness depends on the structure of the ID feature space. As GROOD relies on geometric separation of class prototypes (inspired by Neural Collapse~\citep{papyan2020prevalence}), its performance can degrade when ID representations are less well-clustered. Additionally, the choice of OOD prototype impacts this trade-off. For example, using an ``ID-corrupted val'' prototype improves Near-OOD AUROC to 80.27\% (CIFAR-100) and 83.5\% (ImageNet-1k), while maintaining strong Far-OOD scores.

Importantly, many top-performing methods on the OpenOOD leaderboard require access to OOD data (e.g., OE, CIDER). GROOD remains fully post-hoc and training-free, making it more practical for deployment across varied scenarios.

\begin{table*}[ht]
    \caption{Performance comparison (AUROC \%) on ImageNet-1K using different architectures.}
    \label{tab:arch_comparison}
    \centering
    \resizebox{\textwidth}{!}{%
    \begin{tabular}{l|cc|cc} 
        \toprule
         & \multicolumn{2}{c|}{\textbf{ViT-B-16}} & \multicolumn{2}{c}{\textbf{Swin-T}} \\
        \cmidrule(lr){2-3} \cmidrule(lr){4-5} 
        Method & Near-OOD (\%) $\uparrow$ & Far-OOD (\%) $\uparrow$ & Near-OOD (\%) $\uparrow$ & Far-OOD (\%) $\uparrow$ \\
        \midrule
        \textbf{GROOD (OURS)} & \textbf{76.47\%} & \textbf{90.84\%} & \textbf{76.10\%} & 88.90\% \\
        ReACT \cite{sun2021tone} & 69.26\% & 85.69\% & 75.64\% & 88.23\% \\
        GradNorm \cite{huang2021importance} & 39.28\% & 41.75\% & 47.58\% & 35.47\% \\
        KNN \cite{sun2022knnood} & 74.11\% & 90.60\% & 71.62\% & \textbf{89.37\%} \\
        ASH \cite{djurisic2023extremely} & 53.21\% & 51.56\% & 46.47\% & 44.64\% \\
        \bottomrule
    \end{tabular}
    }  
\end{table*}

\paragraph{Performance on Transformer Architectures}
Further demonstrating the robustness and generalization capabilities of our approach, \cref{tab:arch_comparison} presents the OOD detection performance on ImageNet-1K using Transformer-based architectures, ViT-B-16~\citep{vit} and Swin-T~\citep{liu2021swin}. \ac{GROOD} maintains strong performance, achieving the top AUROC scores for both Near-OOD and Far-OOD detection on ViT-B-16~\citep{vit}, and the best Near-OOD score on Swin-T~\citep{liu2021swin} while being highly competitive for Far-OOD. 
This contrasts significantly with several other methods, such as GradNorm and ASH, whose performance severely degrades on these Transformer architectures compared to their ResNet-based results. 
This suggests that \ac{GROOD}'s mechanism, relying on gradient sensitivity relative to class and OOD prototypes, generalizes more effectively across fundamentally different architectural paradigms than methods potentially more sensitive to specific CNN feature properties.

\subsection{Ablation Studies} \label{subsec:ablations}

To understand the contribution of each component in GROOD, we conduct a series of ablation studies on CIFAR-10 (\cref{tab:ablation_loss}). Each study isolates a key design choice and quantifies its impact on both near-OOD and far-OOD detection performance.

\begin{table}[htbp]
\centering
\caption{Ablation study using different losses and \ac{OOD} scores to show the effectiveness of each proposed part. Evaluation done on CIFAR-10.}
\label{tab:ablation_loss}

\resizebox{0.7\columnwidth}{!}{%
\begin{tabular}{lcc}
\toprule
\multirow{2}{*}{\textbf{Model Variant}} & \multicolumn{2}{c}{\textbf{AUROC (\%)}} \\
 & \textbf{Far-\ac{OOD}} & \textbf{Near-\ac{OOD}} \\
\midrule
(1) Distance to the Noise prototype & 84.3\textsubscript{(\textpm 6.1)} & 79.9\textsubscript{(\textpm 6.5)} \\
(2) Gradient L1-norm only & 92.4\textsubscript{(\textpm 0.48)} & 89.35\textsubscript{(\textpm 0.41)} \\
(3) Grads. wrt class prototypes & 92.7\textsubscript{(\textpm 0.15)} & 89.9\textsubscript{(\textpm 0.05)} \\
(4) OOD prototype with uniform noise only & 91.7\textsubscript{(\textpm 0.6)}  & 88.1\textsubscript{(\textpm 0.55)} \\
\textbf{\ac{GROOD}} & \textbf{93.8}\textsubscript{(\textpm 0.02)} & \textbf{91.16}\textsubscript{(\textpm 0.001)}  \\
\bottomrule
\end{tabular}
}
\end{table}

\paragraph{Distance vs. Gradient-based Scoring} 
Our first experiment examines whether the gradient computation is truly necessary. We compare directly using the distance to the OOD prototype against our full gradient-based approach. The significant performance gap  (79.9\% vs. 91.16\% for near-OOD and 84.3\% vs 93.8\% for far-OOD) demonstrates that gradients capture richer information about sample distribution than raw distances alone.

\paragraph{Nearest Neighbor vs. Gradient Norm}
While prior work like GradNorm~\cite{huang2021importance} uses L1-norm of gradients as the OOD score, we hypothesized that our full approach, \ac{GROOD}, would be more informative. The results support this: \ac{GROOD} achieves 93.8\% AUROC on far-OOD detection and 91.16\% on near-OOD detection compared to 92.4\% and 89.35\% respectively with gradient norm alone, suggesting that the combination of our design choices in \ac{GROOD} provides valuable signal beyond the magnitude of the gradient alone.

\paragraph{OOD vs. Class Prototypes}
A natural question is whether we need a dedicated OOD prototype at all - could we achieve similar results using gradients with respect to class prototypes? The experiment shows that OOD-specific prototypes provide superior performance (93.8\% vs. 92.7\% on far-OOD and 91.16\% vs 89.9\% on near-OOD), validating our design choice to explicitly model out-of-distribution behavior.

\paragraph{Impact of Noise Sources} 
Finally, we investigate the value of our synthetic OOD data generation for \ac{OOD} prototype construction compared to simply using uniform noise. Using only uniform noise degrades performance by 2.1\% on far-OOD (93.8\% vs 91.7\%) and 3.06\% on near-OOD detection (91.16\% vs 88.1\%), respectively, demonstrating the benefit of our comprehensive synthetic approach to prototype construction over basic uniform noise.

These ablations collectively validate \ac{GROOD}'s key design choices: each component contributes meaningfully to the final performance, with the full method achieving the best results across all metrics. The consistent improvements and low standard deviations ($\leq0.6\%$) across experiments indicate the robustness of our approach. 

\subsection{Robustness to Checkpoint Choice}

\begin{table}[!ht]
    \centering
    \caption{Standard Deviation of AUROC for the Last 15 Epochs on CIFAR-100}
    \begin{tabular}{lcc}
        \toprule
        Method & Std (Far OOD) & Std (Near OOD) \\
        \midrule  
        MDS \cite{mahananobis18nips} & 1.05 & 0.24 \\
        ODIN \cite{odin18iclr} & 0.40 & 0.22 \\
        GradNorm \cite{huang2021importance} & 0.62 & 0.72 \\
        VIM \cite{haoqi2022vim}  & 0.95 & 0.37 \\
        \ac{GROOD} & \textbf{0.38} & \textbf{0.2} \\
        \bottomrule
    \end{tabular}
    \label{tab:ood_std}
\end{table}

\begin{figure}[htb!]
    \centering
    \begin{subfigure}{0.45\textwidth}
        \includegraphics[width=\linewidth]{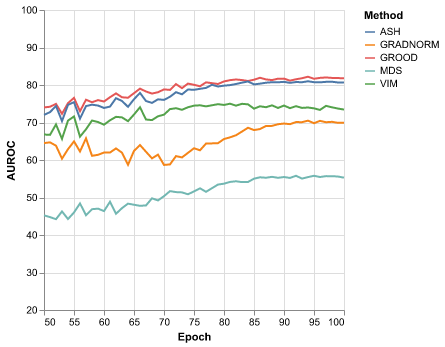}
        \caption{Near-OOD AUROC}
        \label{fig:cifar100_nearood}
    \end{subfigure}
    \hfill
    \begin{subfigure}{0.45\textwidth}
        \includegraphics[width=\linewidth]{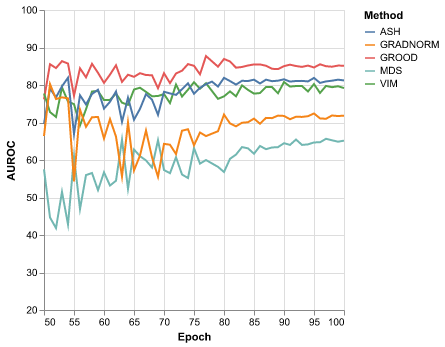}
        \caption{Far-OOD AUROC}
        \label{fig:cifar100_farood}
    \end{subfigure}
    \caption{AUROC Performance on Cifar100 (Near and Far OOD) across different checkpoints showing the stability of GROOD}
    \label{fig:cifar100_ood_ckpt}
\end{figure}

The AUROC metric, while widely used for evaluating \ac{OOD} detection, can exhibit instability during training, particularly in the later stages. This instability means that small fluctuations in the model's weights can lead to significant variations in AUROC scores, making the selection of an optimal checkpoint challenging.
The AUROC curve can vary sharply, even when the test error is relatively stable, indicating a sensitivity to minor weight perturbations. 
In contrast, GROOD's design contributes to more stable \ac{OOD} detection performance. The key intuition behind \ac{GROOD}'s robustness lies in its focus on the sensitivity of weights relative to the \ac{OOD} prototype. 
Throughout training, while the representation space and the \ac{OOD} prototype's absolute location change as the network's weights are updated, their inherent relationship the sensitivity remains stable, leading to consistent \ac{OOD} detection , as further evidenced by the reduced standard deviations shown in \cref{tab:ood_std} and \cref{fig:cifar100_ood_ckpt}.

Further ablation experiments and detailed analysis are provided in the Appendix.

\section{Discussion and Conclusion}

In this work, we introduced GRadient-aware Out-Of-Distribution detection (\ac{GROOD}), a novel approach that combines gradient information with distance metrics to improve detection of \ac{OOD} samples in \ac{DNN}-based image classifiers. Extensive experiments across benchmarks show that \ac{GROOD} effectively detects both near and far \ac{OOD} samples, performs robustly across architectures and datasets, and requires little hyper-parameter tuning, making it practical for deployment in real-world settings~\citep{litjens2017survey, bojarski2016end}.

\ac{GROOD} is particularly well-suited to scenarios where in-distribution features form clear clusters, consistent with the Neural Collapse phenomenon~\citep{papyan2020prevalence}. In such cases, deviations from prototypes are easily detected. However, its performance may degrade when in-distribution data is noisy or poorly separated.

From a practical perspective, \ac{GROOD} improves inference speed compared to KNN-based methods (see \cref{ap:inference_speed}), but storing and processing gradients can be costly for very large datasets or high-resolution images. Prototype construction also matters: while relatively robust, performance can depend on the diversity of auxiliary \ac{OOD} samples (\cref{ap:choice_xood}).

Finally, several design choices proved important. Proximity-based filtering of synthetic prototypes (\cref{sec:practical_proto}) enhanced discriminative power, and mixup toward the \textbf{second}-highest predicted class yielded stronger results than random interpolation, likely because it produces harder boundary samples.

In summary, \ac{GROOD} offers a simple and effective framework for \ac{OOD} detection, especially when in-distribution data is well-structured. Future work could extend it to noisier domains and explore ways to further reduce computational overhead.

\subsubsection*{Acknowledgments}
This work was carried out within the DEEL project. 
It is supported by the DEEL Project CRDPJ 537462-18 funded by the Natural Sciences and Engineering Research Council of Canada (NSERC) and the Consortium for Research and Innovation in Aerospace in Québec (CRIAQ), together with its industrial partners Thales Canada inc, Bell Textron Canada Limited, CAE inc and Bombardier inc.\footnote{\url{https://deel.quebec}}.
The DEEL project is also part of IRT Saint Exupéry and the ANITI AI cluster\footnote{\url{https://www.deel.ai/}}. The authors acknowledge the financial support from DEEL's Industrial and Academic Members and the France 2030 program – Grant agreements n°ANR-10-AIRT-01 and n°ANR-23-IACL-0002.
We are grateful for the digital alliance of Canada and NVIDIA for compute infrastructure. 
We would also like to thank Dr. Samer Nashed for his fruitful discussions and valuable feedback during the review process.

\bibliography{references}
\bibliographystyle{tmlr}

\appendix

\section{Appendix}

\subsection{GROOD Algorithm}

For a comprehensive overview, the complete \ac{GROOD} algorithm is outlined in \cref{alg:GROODinit} and \cref{alg:GROODscore}.
\begin{algorithm}[!ht]
\caption{GROOD initialization: Compute prototypes and gradients for the training set}
\label{alg:GROODinit}
\begin{algorithmic}[1]
\Require Training set $\mathcal{D}_{\text{in}}$, trained model $f$
\Require mixup parameter $\lambda$ 

\State Compute class prototypes $\mathbf{P}_{\text{early}}$ and $\mathbf{P}_{\text{pen}}$  \Comment \cref{eq:proto_compute_class}
\State Compute OOD prototype $\pnoise$ using synthetic data generation\cref{par:synthetic_data} \Comment \cref{eq:proto_compute_ood}

\Function{comp\_grad}{$h, \Ppen, \pnoise$} \label{func:grad}
\State compute $\nabla H(h)$ \Comment \cref{eq:grad} using $ \Ppen$, $\pnoise$ 
\State \Return $\nabla H(h)$
\EndFunction

\For{each $x \in \mathcal{D}_{\text{in}}$}
\State $\nabla (x)=$   \textsc{comp\_grad} $(h(x), \Ppen, \pnoise)$
\EndFor
\State \Return $\{ \nabla (x)\}_{x \in \mathcal{D}_{\text{in}}}$, $\Pearly$, $\Ppen$, $\pnoise$
\end{algorithmic}
\end{algorithm}

\begin{algorithm}[!ht]
\caption{OOD score using GROOD}
\label{alg:GROODscore}
\begin{algorithmic}[1]
\Require Training dataset $\mathcal{D}_{\text{in}}$, trained model $f$
\Require $\{ \nabla(x))\}_{x \in \mathcal{D}_{\text{in}}}$, $\Ppen$, $\pnoise$ from GROOD initialization 
\Require function \textsc{comp\_grad} (in Alg. \ref{alg:GROODinit})
\Require Sample $x_\text{new}$, threshold $\tau$

\State $\nabla (x_\text{new}) =$  $\textsc{comp\_grad}(h(x_\text{new}), \Ppen, \pnoise$)  
\State Compute \ac{OOD} score using Nearest Neighbor search: $S(x_\text{new}) = \min_{x \in \mathcal{D}_{\text{in}}} \| \nabla(x_\text{new}) - \nabla (x) \|_2$ 
\State \textbf{return} \ac{ID} if $S(x_\text{new}) \leq \tau$ else \ac{OOD}
\end{algorithmic}
\end{algorithm}

\subsection{Choice of OOD data for OOD prototype computation} \label{ap:choice_xood}

\begin{table*}[ht]
    \caption{OOD Detection Performance (AUROC \%) by Dataset and OOD Prototype Construction Method.}
    \label{tab:ablate_different_ood}
    \centering
    \resizebox{\textwidth}{!}{%
        \begin{tabular}{l|cc|cc|cc|cc}
            \toprule
            & \multicolumn{2}{c|}{\textbf{Cifar-10}} & \multicolumn{2}{c|}{\textbf{Cifar-100}} & \multicolumn{2}{c|}{\textbf{ImageNet-200}} & \multicolumn{2}{c}{\textbf{ImageNet}} \\
            \cmidrule(lr){2-3} \cmidrule(lr){4-5} \cmidrule(lr){6-7} \cmidrule(lr){8-9}
            OOD Prototype & Near-OOD & Far-OOD & Near-OOD & Far-OOD & Near-OOD & Far-OOD & Near-OOD & Far-OOD \\
            \midrule 
            Synthetic OOD & 91.16 & 93.8 & 78.9 & 84.44 & 83.4 & 92.19 & 78.91 & 94.8 \\
            ID-Corrupted Val & 90.55 & 93.88 & 80.27 & 81.41 & 83.9 & 92.58 & 83.5 & 94.6 \\
            OpenOOD Val & 91.01 & 94.18 & 80.92 & 80.7 & 81.86 & 94.77 & 78.05 & 96.16 \\
            Uniform &   90.9 & 94.1 & 77.26 & 84.5 & 82.84 & 94.46 & 75.23 & 94.55 \\
            Mean of Prototypes & 88.5 & 91.4 & 77.2 & 81.4 &  82.3 & 92.9 & 71.25 & 82.21 \\
            \bottomrule
        \end{tabular}
    }
\end{table*}

\paragraph{OpenOOD Val} To form $\pnoise$, we selected 100 data points from an auxiliary \ac{OOD} validation dataset, as per the OpenOOD framework~\cite{zhang2023openood, yang2022openood}, ensuring no category overlap with test set images. This selection criterion aligns with practices in established post-hoc analyses~\cite{lee2022gradient, mahananobis18nips, kong2021opengan}. Our method demonstrated robustness to the specific choice of \ac{OOD} samples. An investigation with five distinct sets of 100 OOD validation samples each revealed negligible variation in AUROC, with a maximum standard deviation of \(\textbf{0.5\%}\), underscoring our approach's stability across different OOD selections.

\paragraph{ID-Corrupted Val} We further validated our approach using 100 i.i.d. samples from CIFAR-10-C~\cite{hendrycks2019benchmarking} for CIFAR-10 as \ac{ID} and CIFAR-100C for rest of \ac{ID} datasets including CIFAR-100, ImageNet-200 and ImageNet to ensure no possible overlap or leaks to the test set.

\paragraph{Uniform} We try to approximate the representation of \ac{OOD} data by leveraging uniform noise data. Initially, a batch of random noise images is created using uniformly distributed pixel values across all channels. These noise images are then passed through a neural network to extract logits and features from intermediate layers. An energy score is computed for each image, where lower scores indicate a higher likelihood of being \ac{OOD}. The images with the lowest energy scores, which are most similar to out-of-distribution (OOD) data, are selected, and their penultimate layer features are extracted. These features are then aggregated to form an OOD prototype.

\paragraph{Synthetic OOD}  To simulate \ac{OOD} data representations, we utilize a manifold mixup technique on the early layer, similar to the targeted mixup approach described in \cref{sec:proto_computation}. However, our method differs in the interpolation target. Instead of interpolating towards the predicted class prototype, we interpolate towards the second-highest predicted class \(c_2\), which is the closest incorrect class on the decision boundary. 

\paragraph{Mean of Prototypes} Instead of trying to approximate the representation of \ac{OOD} data using auxiliary \ac{OOD} data we rely on the mean of \ac{ID} prototypes.

\Cref{tab:ablate_different_ood} illustrates our method's robustness to different validation \ac{OOD} datasets.

\subsection{Density Plots}
\begin{figure*}[!ht]
    \centering
    \begin{subfigure}[b]{0.45\textwidth}
        \includegraphics[width=\textwidth]{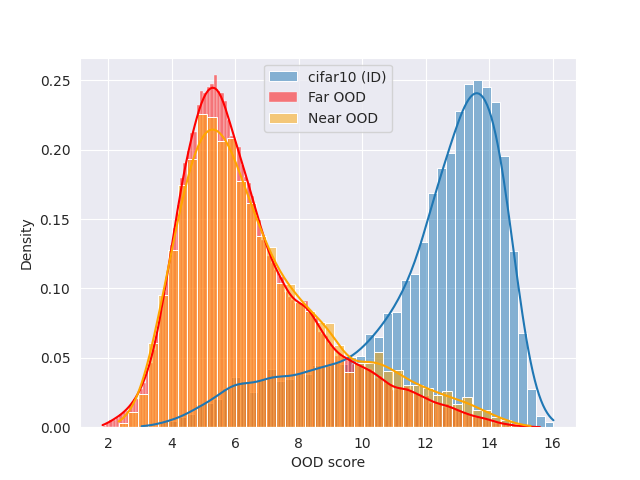}
        \caption{CIFAR-10}
        \label{fig:id_ood_cifar10_combined}
    \end{subfigure}
    \hfill
    \begin{subfigure}[b]{0.45\textwidth}
        \includegraphics[width=\textwidth]{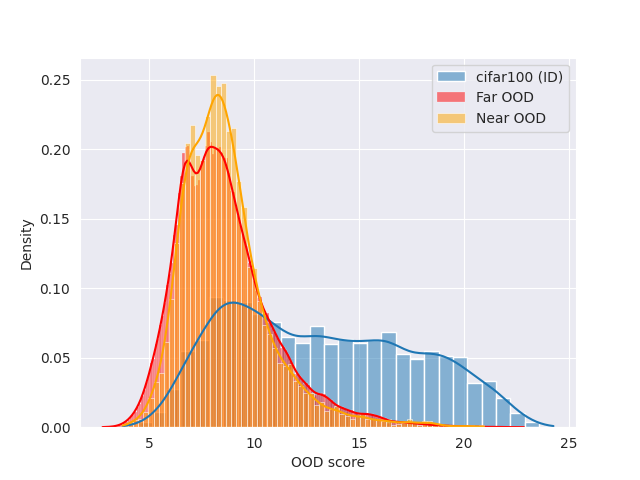}
        \caption{CIFAR-100}
        \label{fig:id_ood_cifar100_combined}
    \end{subfigure}
    \vfill
    \begin{subfigure}[b]{0.45\textwidth}
        \includegraphics[width=\textwidth]{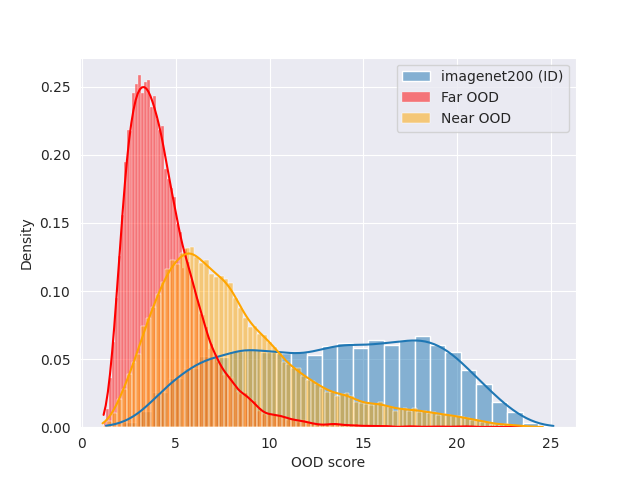}
        \caption{ImageNet-200}
        \label{fig:id_ood_imagenet200_combined}
    \end{subfigure}
    \hfill
    \begin{subfigure}[b]{0.45\textwidth}
        \includegraphics[width=\textwidth]{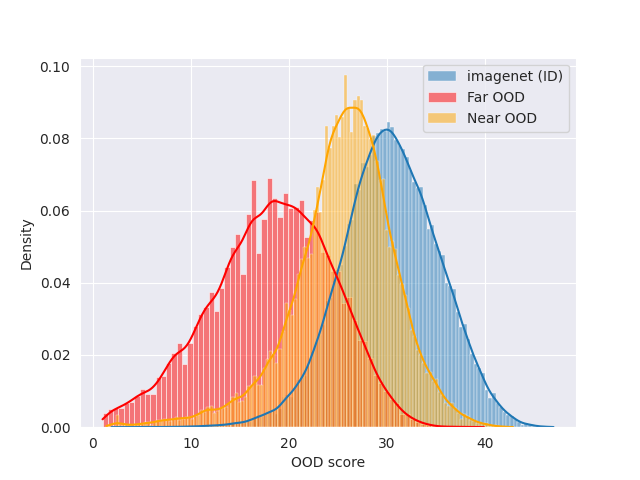}
        \caption{ImageNet-1k}
        \label{fig:id_ood_imagenet1k_combined}
    \end{subfigure}
    \caption{Distribution of \ac{OOD} scores on Near-\ac{OOD} and Far-\ac{OOD} across different datasets.}
    \label{fig:combined_id_ood}
\end{figure*}

To comprehensively evaluate the GROOD method's ability to distinguish between in-distribution (ID) and out-of-distribution (OOD) data, we visualize the distribution of \ac{OOD} scores across a range of datasets with varying characteristics. \Cref{fig:combined_id_ood} presents these visualizations for ID, Near-\ac{OOD}, and Far-\ac{OOD} samples on CIFAR-10, CIFAR-100 (datasets with relatively small, natural images), ImageNet-200 (a subset of ImageNet), and ImageNet-1k (a large-scale, complex dataset). Analyzing performance across this spectrum demonstrates GROOD's robustness to differences in image complexity and dataset size. In each subplot, we use density plots to represent the distribution of \ac{OOD} scores, allowing for a clear visual comparison of separation between ID and OOD distributions.

\subsection{Inference Speed} \label{ap:inference_speed}

GROOD introduces a novel Out-of-Distribution (OOD) detection method involving two forward passes for the mixup part which is inexpensive to compute, a backward pass over the OOD prototype which can be computed using its closed form expression as in \cref{eq:grad} of the main paper and the ne
arest neighbor search which is more computationally intensive. For the latter, Our approach employs the FAISS IndexIVF method for efficient distance computation, utilizing centroids and inverted lists instead of the complete dataset. This technique notably enhances inference speed compared to KNN, particularly in our CIFAR benchmarks. Specifically, on CIFAR-10 and CIFAR-100 datasets, GROOD recorded evaluation inference times over all OOD test sets of \textbf{130} seconds and \textbf{155} seconds, respectively. This is significantly faster than KNN, which took 434 seconds for CIFAR-10 and 641 seconds for CIFAR-100, demonstrating the efficiency of our approach.

\subsection{Ablate value of kth nearest neighbor} 
\begin{figure}[htpb]
  \centering
   \includegraphics[width=0.9\linewidth]{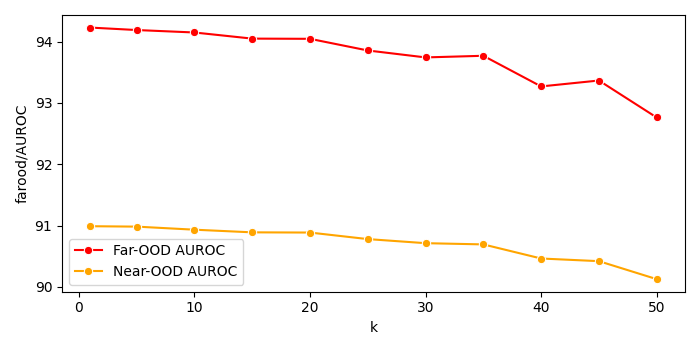} 
  \caption{Ablating different values for k-th nearest neighbor parameter on CIFAR-10}
  \label{fig:ablate_k_cifar10}
\end{figure} 

\Cref{fig:ablate_k_cifar10} illustrates the impact of using the distance to the \( k \)-th nearest neighbor, as proposed by \citet{sun2022knnood}. The plot demonstrates that employing the distance to the nearest point in gradient space leads to optimal results.

\subsection{Gradient computation in closed form} \label{ap:grood_derivation}
This section details the derivation of the closed-form expression for the gradient, as presented in \cref{eq:grad} of the main paper. 
The cross-entropy loss for a logit $L=(L_j)_{j=1}^C+1$ and some (any) \ac{ID} label $y$ is given by:
\[
H(L, y)= - \ln \frac{\exp(L_{y})}{\sum_{i=1}^{C+1}\exp(L_j)}
\]
The partial derivative of this loss with respect to $L_{C+1}$ is given by:
\[
\frac{\partial}{\partial L_{C+1}}H(L,C+1)=\frac{\exp(L_{C+1})}{\sum_{i=1}^{C+1}\exp(L_j)}
\]
which is equal to the softmax probability corresponding to the \ac{OOD} class and does not depend on the specific \ac{ID} class label $y$ anymore.
Now for a feature vector $h$, the corresponding logit vector $L(h)$ is given by \cref{eq:logits}. Since $L_{C+1}(h)=\|h-\pnoise\|_2$ is the only logit depending on $\pnoise$, the gradient of the above loss with respect to $\pnoise$ is given by the chain rule:
\[
\nabla_{\pnoise} H(L(h), y)= \frac{\partial}{\partial L} H(L(h),y)  \nabla_{\pnoise} L(h) =  \frac{\exp(L_{C+1})}{\sum_{i=1}^{C+1}\exp(L_j)} \frac{h-\pnoise}{\|h-\pnoise\|_2}=p_\text{ood}(h)  \frac{h-\pnoise}{\|h-\pnoise\|_2},
\]
as desired.

\subsection{Impact of Mixup-Trained Backbones on GROOD}
\label{ap:mixup-backbone}

We investigate how GROOD's OOD prototype construction interacts with backbones trained using manifold mixup~\citep{verma2019manifold}. Since these models are explicitly trained to generalize across interpolated samples, we hypothesized that synthetic mixup-based OOD prototypes might no longer serve as an effective deviation reference.

\begin{table}[!ht]
\centering
\begin{tabular}{lcc}
\toprule
Method & Near-OOD AUROC (\%) & Far-OOD AUROC (\%) \\
\midrule
GROOD (mean prototype) & \textbf{81.05} & \textbf{80.26} \\
ASH~\citep{djurisic2023extremely} & 79.1 & 56.0 \\
KNN~\citep{sun2022knnood} & 78.0 & \textbf{81.85} \\
GradNorm~\citep{huang2021importance} & 50.0 & 50.0 \\
\bottomrule
\end{tabular}
\caption{Performance of GROOD with mean prototype on a manifold mixup-trained ResNet-18 backbone (CIFAR-100).}
\end{table}

To test this, we evaluated GROOD on a ResNet-18 trained with manifold mixup. Instead of using mixup-based OOD prototypes since it will no longer represent OOD data, we used a mean prototype computed from ID class prototypes. The results are shown below:

These results show that GROOD maintains competitive performance by adjusting its prototype strategy to the model's training procedure. The mixup-based prototype remains optimal for standard-trained models, while alternative strategies like mean prototypes are preferable when the backbone is mixup-regularized.

\end{document}